# Automatic Target Recognition on Synthetic Aperture Radar Imagery: A Survey

O. Kechagias-Stamatis, and N. Aouf

*Abstract*—**Automatic Target Recognition (ATR) for military applications is one of the core processes towards enhancing intelligence and autonomously operating military platforms. Spurred by this and given that Synthetic Aperture Radar (SAR) presents several advantages over its counterpart data domains, this paper surveys and assesses current SAR ATR algorithms that employ the most popular dataset for the SAR domain, namely the Moving and Stationary Target Acquisition and Recognition (MSTAR) dataset. Specifically, we perform a direct comparison between current SAR ATR methods and highlight the strengths and weaknesses of each technique under both standard and extended operational conditions. Additionally, despite MSTAR being the standard SAR ATR benchmarking dataset we also highlight its weaknesses and suggest future research directions.**

## I. INTRODUCTION

AUTOMATIC Target Recognition (ATR) for modern military applications has become a necessity to enhance intelligence, reduce collateral damage and fratricide, and ultimately to support autonomously operating platforms. Thus, for the last decades academia and industrial stakeholders have made various ATR attempts throughout several data domains including 2D Infrared (IR- thermal) [1], [2], 3D Light Detection and Ranging (LIDAR) [3]–[5], 2D Synthetic Aperture Radar (SAR) [6], [7], [16]–[21], [8]–[15] or hybrid 3D ATR originating from 2D visual data [22]. The strengths and weaknesses of each data modality are presented in Table I. Ideally, the sensor and thus the employed data domain, should combine all the positive features of Table I, with the most important ones to be compact, low cost, passive, operating from a long distance, and operating under almost all-weather night-and-day conditions, i.e. 24/7 conditions. However, the latter attribute is a game-changing advantage during military operations, and thus Synthetic Aperture Radar (SAR) SAR is overall an appealing ATR option as it combines the 24/7 advantage, a low ATR processing time, and a long operating range.

Spurred by the advantages of SAR, ATR employing this data domain has been attempted using various methods that either originate from the computer vision domain or techniques that are specifically designed for SAR data aiming at maximizing the target recognition performance. Examples of such methods are employing features, i.e. descriptive encodings of local regions in the SAR imagery, deep neural networks, remapping the SAR imagery in various manifolds and then apply a target recognition strategy, etc. Additionally, SAR ATR algorithms have been evaluated on several datasets including ground-based targets or vessels. However, literature mostly concerns ground targets [23]–[25] with the vast majority employing the public release of the Moving and Stationary Target Acquisition and Recognition (MSTAR) dataset [24]. Indeed, the MSTAR dataset is the most cited SAR ATR dataset with a continuously growing citation progression over the last years (Fig. 1).

Few surveys of SAR ATR are available, such as [26]–[32]. However, our paper is the first to focus on SAR ATR regarding the popular MSTAR dataset. Besides, our paper comprehensively covers different ATR schemes highlighting the various methods employed. Finally, our work extends the existing survey-type of literature with the major contributions of this work summarized as follows:

a. To the best of our knowledge, this is the first survey paper to comprehensively cover state-of-the-art and most recent SAR ATR methods that spread over a range of various disciples.

b. Opposed to existing surveys [26]–[32], we specifically focus on the MSTAR dataset rather than presenting solutions involving other SAR datasets. This is important as the MSTAR has established itself as the major benchmarking dataset in the SAR ATR domain.

c. Despite the numerous papers exploiting the MSTAR database (see Fig. 1), this paper covers the most recent and advanced progress of SAR ATR. Therefore, it provides the reader with state-of-the-art methods.

d. Comprehensive comparisons of existing ATR algorithms with summaries, performance assessments, and future research

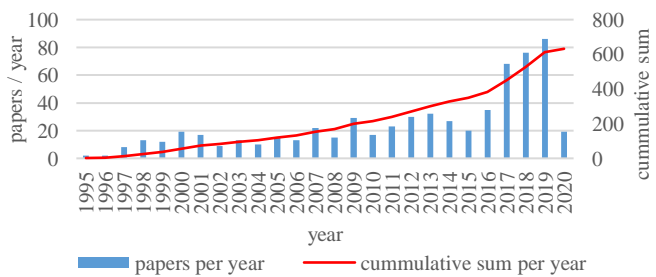

**Fig. 1.** MSTAR citation progression till mid 2020

O. Kechagias-Stamatis and N. Aouf are with the Department of Electrical and Electronic Engineering, City University of London, EC1V 0HB, UK (e-mail: odysseas.kechagiasstamatis@city.ac.uk).

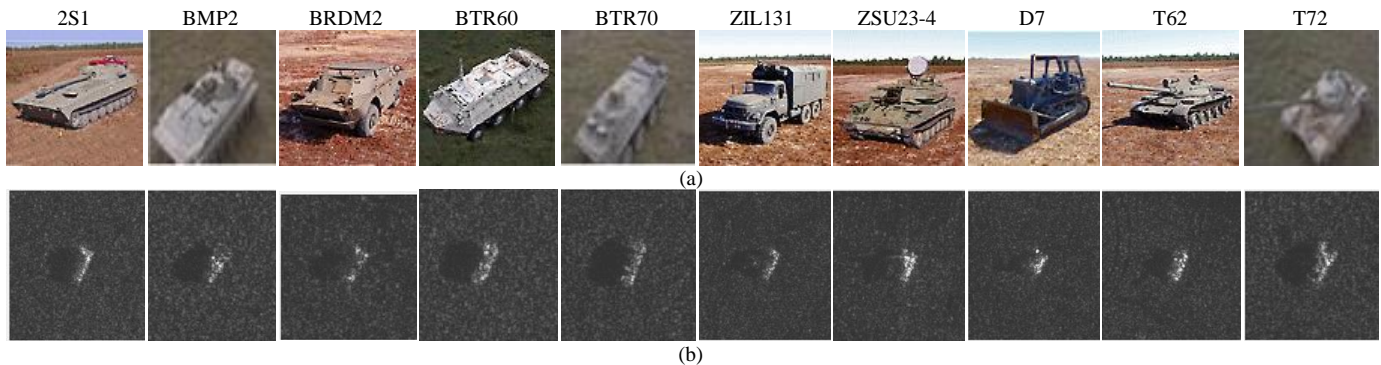

Fig. 2. Examples of standard MSTAR targets (a) visual image (b) SAR image

TABLE II
STANDARD MSTAR DATABASE

| Target | BMP2 | | | BTR70 | T72 | | | BTR60 | 2S1 | BRDM2 | D7 | T62 | ZIL131 | ZSU23-4 |
|---|---|---|---|---|---|---|---|---|---|---|---|---|---|---|
| ID | 1 | 2 | 3 | 4 | 5 | 6 | 7 | 8 | 9 | 10 | 11 | 12 | 13 | 14 |
| serial N° | 9563 | 9566 | c21 | c71 | 132 | 812 | s7 | k10yt7532 | b01 | e71 | - | a51 | e12 | d08 |
| train 17° | 233 | 232 | 233 | 233 | 232 | 231 | 228 | 256 | 299 | 298 | 299 | 299 | 299 | 299 |
| test 15° | 195 | 196 | 196 | 196 | 196 | 195 | 191 | 195 | 274 | 274 | 274 | 273 | 274 | 274 |
| test 30° | - | - | - | - | - | - | - | - | 288 | 287 | - | - | - | 288 |
| test 45° | - | - | - | - | - | - | - | - | 288 | 287 | - | - | - | 303 |

directions are also presented.

The structure of this paper is as follows. Section 2 presents the MSTAR dataset, while Section 3 the various SAR ATR methods. Section 4 compares and contrasts the performance of the presented ATR algorithms, while Section 5 concludes this work and proposed future research directions.

## II. MOVING AND STATIONARY TARGET ACQUISITION AND RECOGNITION DATASET

The MSTAR dataset comprises of X-band SAR imagery with a 1x1 ft resolution, 360° articulation with a 1° spacing, and an image size of 128x128 pixels [32]. The commonly used targets in the literature are presented in Table II and Fig. 2, while for a complete list of all target variants including the SLICY static target, the reader is referred to [23]. It should be noted that despite the MSTAR dataset offers SLICY, the majority of the literature does not involve it during SAR ATR evaluation and thus in this work it will be omitted.

The proposed SAR ATR methods are evaluated on two major target configuration sets namely the Standard Operation Conditions (SOC) and the Extended Operation Conditions (EOC). The former typically includes the 10 target classes of Table II or a subset comprising of three targets. EOC involves the same 10-class targets but with various acquisition geometries, i.e. depression angle variations up to 45°, target state variations, i.e. articulation, , alternative configurations, and intra-class variability, and finally local target deployment, i.e. various obscuration levels [33], [34].

TABLE I
ANALYSIS OF DATA DOMAIN-DRIVEN ATR SOLUTIONS

| | 3D LIDAR | 2D visual | 2D IR | 2D SAR |
|---|---|---|---|---|
| Template size | single 3D model | multiple views | multiple views | multiple views |
| Target pose invariance | + | - | - | - |
| Target illumination invariance | + | - | + | + |
| ATR based on the underlying structure | + | - | - | + |
| Operating day and night | + | - | + | + |
| ATR Processing time | - | + | + | + |
| Equipment cost | - | + | + | - |
| Equipment size | - | + | + | - |
| Power consumption | - | + | + | - |
| Data acquisition rate | - (scanning LIDAR) + (flash LIDAR) | + | + | - |
| Maximum operating range | ≈100m | >100m | <100m | >>100m |
| Reveal sensor position | active | passive | passive | active |

However, despite the MSTAR dataset has established itself as the major SAR ATR benchmarking dataset, it is worth noting that it suffers from various deficiencies that can be grouped into dataset-linked and user-linked:

a. *Background correlation.* This is one of the major dataset-linked flaws and is related to the strong background correlation among the SAR images allowing high true positive rates even if the target is absent from the scene [35], [36].

b. *Non-standard target patch size.* A number of papers aim to reduce the effectiveness of the background correlation either by cropping the target from the scene using a fixed mask size or by exploiting a target segmentation strategy. However, the final SAR patch size that contains the target is not fixed and thus may still contain background regions affecting the final ATR performance and making a direct comparison between the available ATR methods not trivial.

c. *Non-standard SOC and EOC trials.* Despite [33], [34] define the SOC and EOC target setups, over the years additional SOC and EOC setups have emerged by establishing various inter-class (different classes of targets) and intra-class target configurations (variations of the same target class) from the standard MSTAR targets presented in Table II. However, the multiple SOC and EOC setups prohibit an extended and direct comparison among the proposed SAR ATR algorithms, and evaluations are constrained to common target subsets. In fact, the MSTAR related literature offers more than 17 types of 10-class and 3-class SOC trials, and more than 33 types of EOC.

TABLE III
MOST REFERENCED MSTAR SOC AND EOC EVALUATION SUBSETS

| SOC | SOC ID of most referenced target subsets (involved targets) | | |
|---|---|---|---|
| 10-class | 1 (all target serial N° for training and testing) | | 2 (2S1 - BMP2 (9563) – BTR60– BTR70 – D7 – T62 – T72 (132) - ZIL131 - ZSU23/4 for training and testing) |
| 3-class | 1 (all BMP2 classes – BTR70 – all T72 classes, both for training and testing) | | |

| EOC | Training targets | Testing targets | Description |
|---|---|---|---|
| depression variation | 2S1 - BRDM2 -ZSU23/4 | 2S1 - BRDM2 - ZSU23/4 | Testing at 15°, 30° and 45° depression angles |
| version variants | BMP2 (9563) – T72 (132) – BTR60 – T62 | BMP2 (9566) – BMP2 (c21) – BMP2 (812) – T72 (s7) – BTR60 – T62 | Testing at 15° against training targets and BMP2 version variants |
| noise variation | 2S1 – BMP2 (9563) – BTR60 – BTR70 – D7 – T62 – T72 (132)- ZIL 131 – ZSU23/4 | 10-class with SOC ID 2 | Testing at 15° with additive artificial Gaussian noise at $SNR = \{10, 5, 0, -5, -10\}dB$ |
| occlusion variation | 10-class with SOC ID 2 | 10-class with SOC ID 2 | Testing at 15° with additive occlusion level at $\{10, 20, 30, 40, 50\}\%$ |
| resolution variation | 10-class with SOC ID 2 | 10-class with SOC ID 2 | Testing at 15° with resolution variation at $\{0.3\times0.3, 0.4\times0.4, 0.5\times0.5, 0.6\times0.6, 0.7\times0.7\}m$ |

d. *A priori known training and testing imagery*. Both training and testing target sets are *a priori* known making the evaluation unrealistic and questioning the true effectiveness of the proposed methods against unknown testing sets.

Though, despite these deficiencies, the MSTAR dataset is currently the standard SAR ATR benchmarking dataset with numerous citations (Fig. 1). The most common SOC and EOC configurations presented in the literature are shown in Table III. However, for better readability, the performance comparison of various current techniques presented in Section 5 will involve only the most common MSTAR evaluation subsets presented in Table III.

## III. ENCODING AND CLASSIFYING SAR IMAGERY

SAR image encoding is the first major stage for a SAR ATR system. Conceptually, the simplest approach is to neglect any data domain remapping and directly encode the raw SAR reflectivity. Though, the simplicity of this strategy does not underestimate the complexity of the encoding process itself but rather reflects the immediacy of applying the encoding technique. On the other side, some methods initially transform the SAR data into a different data domain and then encode it. Hence, we define two major encoding schemes, i.e. the reflectivity attribute-based, and the transform-based, accordingly. However, regardless of the encoding scheme and data domain employed, the second major phase of ATR incorporates matching the template encodings against the target one and ultimately classifying the target. Despite the encoding and classification phases are discrete, the SAR ATR methods offered by the literature are presented as a complete algorithm where the two major phases are heavily associated and cross-tuned to gain the maximum ATR performance. Thus, it is more appropriate to present and evaluate each technique as a whole pipeline rather than two isolated schemes. Hence, for better readability, we group the current SAR ATR methods based on the data domain utilized and subgroup them depending on the governing method employed.

### A. Reflectivity Attribute-Based Methods

This type of technique exploits the raw SAR imagery without involving any pre-processing phase to alter the initial data domain. As described in the subsections, this group of methods involves, feature-based methods, scattering centers calculation, low-rank matrix factorization, deep learning, sparse representation classification, and hybrid strategies combining two of the reflectivity methods.

#### 1) Feature-based methods

These methods describe the SAR image by a set of attributes, i.e. handcrafted features, or moments, that are sufficiently descriptive to achieve appealing target classification under various nuisances. An overview of the methods is presented in Table IV.

Belloni *et al.* [36] initially segment the target from the scene to reduce any background correlation effects by employing a Gaussian Mixture Model (GMM) scheme [37]. Then several features are individually used to encode the SAR image, where top performance is attained using the Binary Robust Invariant Scalable Keypoints (BRISK) [38] features. Feature matching is based on a Nearest Neighbor Distance Ratio (NNDR) scheme utilizing the Hamming distance [39]. An example of BRISK features is presented in Fig. 3. Amrani *et al.* [40] generate a graph-based visual saliency map [41] to extract the target from the SAR image, which is then encoded based on Gabor and Histogram of Oriented Gradients (HOG) [42] features. The latter ones are then fused and input to a two-level directed acyclic graph support vector metric learning (DAG-SVML). This strategy aims to reject weak classifiers via the DAG, emphasize the relevant features, and reduce the influence of noninformative features utilizing the Mahalanobis metric.

Bolourchi *et al.* [43] suggest a moment-fusion strategy to describe the input image and rank the moments in a descending order based on their Fisher score, which are then input to an SVM type of classifier. Experiments involve both cartesian and radial moments, however, fusing the Zernike, pseudo-Zernike,

TABLE IV
FEATURE-BASED METHODS

| N° | Reference | SAR chip size | Pre-processing | Main features |
|---|---|---|---|---|
| 1 | Belloni *et al.* [36] | variable | - | BRISK features, Hamming distance for classification |
| 2 | Bolourchi *et al.* [43] | 128x128 | Histogram equalization, averaging filter, thresholding | LibSVM classification |
| 3 | Amrani *et al.* [40] | 128x128 | - | Gabor and HOG features, Mahalanobis distance for classification |
| 4 | Tan *et al.* [45] | 88x88 | Histogram equalization, averaging filter, thresholding, morphological operations | Target outline "dictionary", a variant of Hausdorff distance metric for classification |
| 5 | Clemente *et al.* [46] | 128x128 | - | Krawtchouk moments, k-NN classification |

TABLE V
SCATTERING GEOMETRY INTERPRETATION OF $\alpha$ AND L

| Scattering geometry | $\alpha$ | L |
|---|---|---|
| Trihedral | 1 | 0 |
| Top Hat | 0.5 | 0 |
| Corner diffraction | -1 | 0 |
| Sphere | 0 | 0 |
| Edge broadside | 0 | >0 |
| Edge diffraction | -0.5 | >0 |
| Dihedral | 1 | >0 |
| Cylinder | 0.5 | >0 |

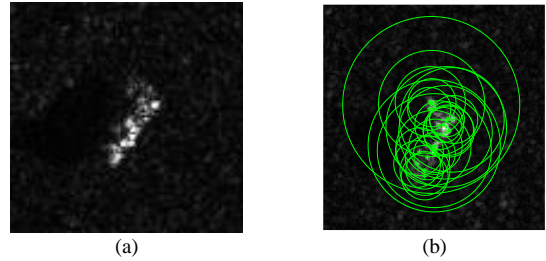

(a)                    (b)

**Fig. 3.** (a) MSTAR SAR image (b) BRISK features (the center of each circle represents the keypoint location with the corresponding radius the scale where the keypoint is detected)

Fourier–Merlin, Chebyshev–Fourier, radial Chebyshev, and radial harmonic Fourier moments afford a competitive ATR performance. It is worth noting that the involved moments oversample the central part of the SAR image where the target is placed, enhancing the overall performance. Finally, classification is based on the Library Support Vector Machine (LibSVM) [44].

Clemente *et al.* [45] extract the discrete-defined Krawtchouk moments, which are then normalized and input to a k-NN classification scheme (k=3). In contrast to current literature that is constrained to purely target classification, this work considers all three ATR phases, i.e. target recognition, identification, and classification

A different strategy that extracts the target outlines is proposed by Tan *et al.* [46]. The latter ones are then segmented into several parts and matched independently with the template outlines to increase the robustness of the proposed method under target deformations. Finally, a variant of the Hausdorff distance metric is used as the similarity measure between the target and the template outline to determine the target's class. Further methods include binary operations [47], sparse robust filters [48], and the azimuth and range target profile fusion [49].

### 2) Attributed Scattering Centers (ASC)

Scattering centers or scatterers are the dominating high-frequency radar reflections from surfaces such as corners and flat plates. ATR techniques that exploit the scattering centers, model the scattering behavior predicted by physical optics and the geometrical theory of diffraction and statistically estimate the object's location, geometry, and polarization response. Each scattering center is a function of frequency $f$ and aspect angle $\varphi$, and the total backscattering field is the sum of these individual scattering centers [50], [51]:

$$E(f, \varphi; \Theta) = \sum_{i=1}^{No} E_i(f, \varphi; \theta_i) \qquad (1)$$

$$
\begin{aligned}
E_i(f, \varphi; \theta_i) = A_i &\times \left( j \frac{f}{f_c} \right)^{\alpha_i} \\
&\times \exp\left[ -j \frac{4\pi f}{c} (x_i \cos\varphi + y_i \sin\varphi) \right] \\
&\times \operatorname{sinc}\left( \frac{2\pi f}{c} L_i \sin\left(\varphi - \overline{\varphi_i}\right) \right) \\
&\times \exp\left( -2\pi f \gamma_i \sin\varphi \right)
\end{aligned}
\qquad (2)
$$

where $f_c$ is the radar center frequency, $c$ is the propagation velocity, $No$ is the number of scatterers with $\Theta = [\theta_1, ....., \theta_i, ....., \theta_{No}]$ incorporating the parameter matrices of these scatterers. $E_i(f, \varphi; \theta_i)$ is the response from the i$^{th}$ scatterer defined by the parameter tensor $\theta_i = [A_i, x_i, y_i, L_i, \overline{\varphi_i}, \alpha_i, \gamma_i]^T$, $A_i$ is the relative amplitude, $\alpha_i \in \{-1, -0.5, 0, 0.5, 1\}$ is the frequency and geometry dependence, and $x_i$ and $y_i$ are the scatterer's location. According to the length $L_i$, a scattering center can be categorized into a localized scattering center ($L_i = 0$) and a distributed scattering center with ($L_i \neq 0$). $\overline{\varphi_i}$ is the orientation angle of distributed scattering center, $\gamma_i$ is the aspect dependence of localized scattering center, $()^T$ denotes the transpose operator. The scattering geometries are distinguishable by their parameters $(\alpha, L)$, examples of which are presented in Table V [52], [53]. In the context of SAR ATR, the scattering centers create a 2D attributed scattering center (ASC) model which is used to characterize the target [54]. An example depicting the ASC model is presented in Fig. 4, while an overview of the methods to be presented is shown in Table VI.

TABLE VI
ATTRIBUTED SCATTERING CENTERS

| Nº | Reference | SAR chip size | Pre-processing | Main features |
|---|---|---|---|---|
| 1 | Li and Du [53] | 64x64 | - | $L_2$-norm minimum reconstruction error for classification |
| 2 | Zhang [63] | 128x128 | - | Hungarian algorithm for classification |
| 3 | Ding et al. [66] | 128x128 | - | D-S theory for classification |
| 4 | Ding et al. [65] | variable | morphological operations | SVM for classification |
| 5 | Ding et al. [60] | 128x128 | - | two-step classification: coarse using Hungarian algorithm and fine using D-S theory |
| 6 | Ding et al. [54] | 128x128 | CFAR, morphological operations, watershed segmentation | Hungarian algorithm for classification |
| 7 | Ding and Wen [64] | 128x128 | - | feature fusion, score-based classification |

Ding et al. [54] propose a SAR ATR algorithm based on the ASC modelling concept [55]. Initially, the SAR response of the target is segmented from the background by combining a Constant False Alarm Rate (CFAR) strategy with various morphological operations. Then the processed image undergoes a multi-staged processing pipeline that involves highest SAR energy segmentation via the watershed algorithm, optimal parameter estimation for the scattering model via a novel Quasi-Newton method [56], error estimation between the reconstructed SAR image based on the ASC and the original SAR image via the CLEAN method [57]. The ASC parameters that present the smallest reconstruction error are the ones used to remap the SAR image into the ASC feature space. Finally, template matching in the ASC feature space is performed via the Hungarian algorithm [58], [59]. Ding et al. [60] extract from the target image the ASC features employing the approximate maximum likelihood (AML) method [61], and create a one-to-one correspondence with the template images using the Hungarian algorithm. These correspondences are then fine matched through an amplitude and a structure similarity using the Dempster–Shafer (D–S) theory of evidence [62]. Accordingly, Zhang [63] also employs ASC features. Ding and Wen [64] fuse global and local features where the former are the SAR image intensities of the entire image and the latter features are extracted from the ASC model. Ding et al. [65] enhance the quality and the completeness of the template dataset by suggesting a multi-level SAR imagery reconstruction using the ASC model. Each target is segmented from the background via a CFAR scheme. On the segmented target, a 9x9 grid is overlaid and the ASC model for each sub-region is individually estimated based on the AML technique. The advantage of this multi-level method is a local ASC parameter setup, rather than a global which is the norm for ASC based SAR ATR. Ding et al. [66] segment a region $R$ that encloses the strongest ASC response utilizing the watershed algorithm. Then the $\theta$ parameter of that segment is iteratively optimized using a novel quasi-Newton method [56] until the ASC model subtracted from the original SAR image produces a residual error below a defined threshold, or when the number of maximum iterations is reached. Li and Du [53] suggest a multi-staged method that initially convolves the ASC model with several filter sizes defined by a genetic algorithm to create a set of low-level feature maps. The latter are sparsely encoded using a proposed locality constraint discriminative dictionary learning (LcLcDDL) that is created during the training phase. Then the low-level local features are converted into high-level global features by employing a spatial pyramid matching scheme.

### 3) Sparse Representation Classification (SRC)

It has been proven that a test image $y$ can be adequately represented by intra-class training templates. Given $X = [X_1, ..., X_k, ..., X_c] \in R^{m \times n}$ the $n$ templates of size $m$, $X_k = [x_{k,1}, x_{k,2}, ..., x_{k,n_k}] \in R^{m \times n_k}$ a template subset containing $n_k$ templates from the k$^{th}$ class with $n = \sum n_k$, the SRC theory assumes that if $y$ belongs to the k$^{th}$ class, it will approximately lie in the linear span of $X_k$, $y \approx X_k a_k$. Where $a_k = [a_{k,1}, a_{k,2}, ..., a_{k,n_k}]^T \in R^{n_k}$ is the coefficient vector with entries being the weight, i.e. atoms, of the corresponding training samples in $X_k$, also known as the *dictionary*.

Given that, the test image class is not *a priori* known SRC assumes that $y$ is a linear combination of the entire dataset $X$, the cardinality of the object classes is large, and the representation of $y$ is naturally sparse, SRC aims at finding the sparsest solution to $y = Xa$, which can be derived by solving the optimization problem

$$a* = \arg \min_a \|y - Xa\|_F^2 + \gamma \|a\|_0 \qquad (3)$$

where $a* = [a_1^*, ..., a_k^*, ..., a_c^*]$, $\gamma \geq 0$ and $\|\cdot\|_0$ is the $l_0 - norm$ counting the non-zero entries in $\alpha$. Provided that the problem in Eq. (3) is NP-hard, compressive sensing theory [67] suggests that if $y$ is sparse enough the initial signal can be recovered by substituting the $l_0 - norm$ with the $l_1 - norm$

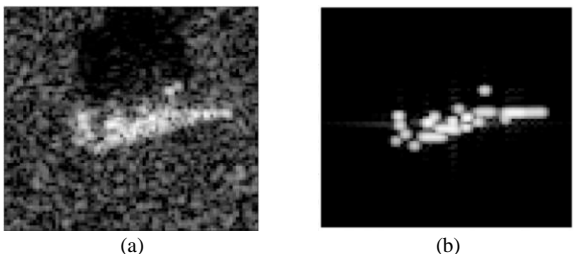

(a)                                    (b)

**Fig. 4.** Target representation using the ASC model (a) original MSTAR image (b) corresponding ASC model

TABLE VII
SPARSE REPRESENTATION CLASSIFICATION

| N⁰ | Reference | SAR chip size | Pre-processing | Data augmentation | Main features |
|----|-----------|---------------|----------------|-------------------|---------------|
| 1 | Wei *et al.* [68] | 128x128 | - | - | $l_{1,\infty} - norm$ |
| 2 | Miao and Shan [70] | 128x128 | - | - | multiple target views (6) |
| 3 | Liu and Chen [71] | 80x80 | normalization | - | $l_2 - norm$ , D-S theory of evidence, maximum belief rule |
| 4 | Ren *et al.* [72] | 64x64 | - | - | Tikhonov regularization matrix, learned dictionary |
| 5 | Ren *et al.* [73] | 64x64 | - | - | weighted regression strategy |
| 6 | Chang and You [74] | 128x128 | normalization, noise filtering, morphological operations | - | two-stream SRC, score-level fusion |
| 7 | Huang *et al.* [75] | 128x128 | - | - | multiple target views (3) |
| 8 | Liu *et al.* [76] | 80x80 | normalization | - | statistical dictionary learning |
| 9 | Zhang [77] | 128x128 | - | multi-resolution | $l_{0,2} - norm$ |
| 10 | Ding and Wen [78] | 128x128 | - | multi-resolution | $l_0 - norm$ , $l_2 - norm$ minimum reconstruction error |
| 11 | Li *et al.* [79] | 80x80 | - | - | analysis dictionary learning |
| 12 | Zhang and Liu [80] | 128x128 | - | multi-resolution | $l_{0,2} - norm$ |
| 13 | Zhang *et al.* [81] | 64x64 | - | - | multiple target views (6) + $l_0 - norm$ |

$$\boldsymbol{a}^* = \arg\min_{\boldsymbol{a}} \|\boldsymbol{y} - \boldsymbol{Xa}\|_F^2 + \gamma \|\boldsymbol{a}\|_1 \tag{4}$$

However, advances in the SRC domain extend to several values within the $l_p - norm$ space $p \in (0, \infty)$ .

The atoms in $\boldsymbol{\alpha}$ are ideally related to samples that belong to the same class as $y$ . Hence, during the SRC classification $y$ is assigned to the class that fulfils the following constraint:

$$class(y) = \arg\min(e_k) = \arg\min\left(\|\boldsymbol{y} - \boldsymbol{X}_k \boldsymbol{a}_k^*\|\right) \tag{5}$$

where $\boldsymbol{a}_k^*$ is a vector with all entries zero except the ones associated with class $k$ and $e_k$ is the reconstruction error. In the context of SAR ATR, the various SRC methods utilize different dictionary constructions and $l_p - norm$ solutions. An overview of the current methods is presented in Table VII.

Wei *et al.* [68] suggest a fast discriminative dictionary learning (FaDDL) method that reduces the execution time without losing the classification accuracy, by adopting an $l_{1,\infty} - norm$ ball as a constraint to replace $l_0 - norm$ and $l_1 - norm$ that are typically used with the discriminative dictionary learning dictionaries. For completeness, it is worth noting that the general formulation of the $l_{1,\infty} - norm$ is; considering the projection of a matrix $A$ to the $l_{1,\infty}$ ball to find a matrix $B$ that solves the convex optimization problem [69]:

$$l_{1,\infty} : \min_{B,\mu} \frac{1}{2} \sum_{i,j} \left( B_{i,j} - A_{i,j} \right)^2 \tag{6}$$

$$
\begin{aligned}
s.t. \quad &\forall i, j \; B_{i,j} \leq \mu_i \\
&\sum_i \mu_i < C \\
&\forall i, j \; B_{i,j} \geq 0 \\
&\forall i \; \mu_i \geq 0
\end{aligned}
\tag{7}
$$

where Eq. (6) is the Euclidean distance between $A$ and $B$, given that $B$ is in the boundary of the $l_{1,\infty}$ ball of radius $C$. Variables $\mu_i$ constrain the coefficients of $B$ as of Eq. (7).

Zhang [70] proposes a multi-stream SRC configuration, where each stream utilizes the target image under different resolution. The SRC scheme employs a $l_{0,2} - norm$ , which is a mixed $l_p - norm$ variant that first applies the $l_0 - norm$ on each row of the dictionary matrix, and then the $l_2 - norm$ on the resulting vector. Ding and Wen [71] also suggest a multi-resolution method but augment the SRC dictionary by creating multiple SAR images of lower resolutions originating from the original templates. Similarly, Zhang and Liu [72] augment the templates with multi-resolution representations, which then undergo a Principal Component Analysis (PCA) prior to the SRC scheme. Liu and Chen [73] weaken the influence of the target aspect angle sensitivity by normalizing the reconstruction error $e_k$ and thus enhance the contribution of the templates with small $e_k$ values. Finally, the normalized errors are fused via the Dempster–Shafer theory [62] and the final target classification relies on the maximum belief rule [74]. Chang and You [75] decouple the three main contributors of the SAR imagery, i.e. target region, shadow, and background, and input the segmented target (target region and shadow) along with the original image into two distinct SRC pipelines. Then both classification results are combined by a score-level fusion for target recognition.

Driven by the conclusion that sparse models learned from training samples have more advantages than the predefined ones [76], [77], Liu *et al.* [78] generate a discriminative statistical dictionary learning (DSDL) technique to enhance the SRC performance. The discriminative nature of the DSDL is further enhanced by manipulating the dictionary entries by minimizing the intra-class differences and maximizing the inter-class differences. Similarly, Ren *et al.* [79] propose a class-oriented dictionary learning scheme. The latter modifies Eq. (4) by adding the Tikhonov regularization matrix $\boldsymbol{\Gamma}$ :

$$\boldsymbol{a}* = \arg\min_{a} \left\| \boldsymbol{y} - \boldsymbol{Xa} \right\|_{2}^{2} + \gamma \left\| \boldsymbol{\Gamma a} \right\|_{1} \qquad (8)$$

$$\boldsymbol{\Gamma} = \begin{bmatrix} d_{i,j}(x_{i,j}, a_{i,1}) & \cdots & 0 \\ \vdots & \ddots & \vdots \\ 0 & \cdots & d_{i,j}(x_{i,j}, a_{i,n_i}) \end{bmatrix} \qquad (9)$$

Li *et al.* [80] suggest a hybrid-structured synthesis dictionary to better reflect the inter-class target differences that incorporate a learned dictionary $\boldsymbol{D}^s$, which is the same for all templates and reduces the effect of common features, such as the high similarity caused by specular reflection. Rather than using the processing costly non-linear $l_0 - norm$ and $l_1 - norm$ sparsity regularizers, the authors employ an analysis dictionary $\boldsymbol{P}$ [81] to create the coding coefficients of the dictionary by linear projection.

Opposing to the previous methods that utilize a single target aspect, several SAR ATR pipelines exploit multiple target aspects. Joint SRC [82] for example, exploits three target views to increase the completeness of the target's SAR signature and then utilizes a mixed $l_{0,2} - norm$. The reasoning for using multiple views is that these are highly correlated sharing the same response pattern within the dictionary and thus this conciseness can enhance the overall ATR performance. Miao and Shan [83] first cluster the SAR templates into several aspect sets based on an image correlation criterion. Then the Joint SRC (JSRC) [84] is applied utilizing six images of the same target but in various views. The JSRC problem is solved employing an $l_{1,2} - norm$ (it first applies the $l_2 - norm$ on each row of the matrix, and then the $l_1 - norm$ on the resulting vector). Later, the reconstruction errors $e_k$ from JSRC (one $e_k$ per target view) are normalized and input to a Bayesian decision fusion scheme where the classification decision relies on the maximum posterior probability. In the event a single target view is available the proposed algorithm degrades to the typical SRC scheme. Ren *at al.* [85] propose a method that learns a supervised sparse model from training single-view samples by utilizing sample label information. Additionally, during dictionary learning, a supervised classifier is jointly designed to enhance the classification performance and the classification error is backpropagated to the dictionary learning process to optimize the dictionary atoms. To enhance the classification performance, during testing a multi-view strategy is applied combined with a weighted regression strategy to assign the identity. Huang *et al.* [86] reduce the classification error imposed by speckle noise by generating a joint low-rank and sparse multi-view denoising (JLSMD) dictionary. Creating the JLSMD dictionary requires combining several target views and exploiting the low-rank property of multi-view target images and the sparsity of speckle-noise for SAR systems. The target identity is assigned based on an $l_1 - norm$ scheme.

*4) Low-rank matrix factorization*

Nonnegative matrix factorization (NMF) and Truncated singular value decomposition (TSVD) are two Low-rank matrix factorization (LMF) methods that are used for SAR ATR. Given $\boldsymbol{X} = [\boldsymbol{x}_1, ..., \boldsymbol{x}_n] \in R^{m \times n}$ that includes $n$ samples and $m$ dimensions, NMF aims to find two non-negative matrices $\boldsymbol{Z}$ and $\boldsymbol{H}$ so that

$$\min_{\boldsymbol{Z} \geq 0, \boldsymbol{H} \geq 0} \left\| \boldsymbol{X} - \boldsymbol{ZH} \right\|_F^2 \qquad (10)$$

where $\boldsymbol{Z} = [\boldsymbol{z}_1, ..., \boldsymbol{z}_d] \in R^{m \times d}$ and $\boldsymbol{H} = [\boldsymbol{h}_1, ..., \boldsymbol{h}_n] \in R^{d \times n}$ are low-rank matrices, with $\boldsymbol{Z}$ mapping the templates $\boldsymbol{X}$ with their low-dimensional representation $\boldsymbol{H}$. For the TSVD case, the factorized matrix $\boldsymbol{Z}$ undergoes an orthogonality constraint and the corresponding optimization problem is defined as

$$\min_{\boldsymbol{Z}^T \boldsymbol{Z} = I_d, \boldsymbol{H}} \left\| \boldsymbol{X} - \boldsymbol{ZH} \right\|_F^2 \qquad (11)$$

Latest LMF techniques are mostly employing an NMF approach [11], with Dang *et al.* [87] suggesting an incremental NMF (INMF) that has sparse $l_p$ constraints, where $p = 1/2$ to employ a variety of norm-sparse constraints. In contrast to the typical NMF where existing and new samples are computed to retrain a new model, the INMF involves only the new samples to update the trained model incrementally. Hence, the proposed $l_p - INMF$ applies an updating process under a generic sparse $l_p$ constraint during the matrix decomposition of the INMF process. Similarly, in [15] the authors suggest the $l_{1/2} - NMF$ strategy where the NMF features are input to an SRC scheme that employs an $l_{1/2}$-norm optimization to identify the sparsest solution. Zhang *et al.* [88] first extract three types of features, i.e. Gabor, PCA, and wavelet features, which then undergo an NMF process to enhance the modelling data sparsity. Finally, the target and the template NMF-processed features are then input to a two-staged classification process that includes a coarse classification phase based on a voting process, and a fine classification phase relying on a Bayesian decision fusion process.

TSVD has also been employed with Yu *et al.* [89] augmenting the typical TSVD optimization (Eq. 11) with a multi-manifold regularization term to achieve better ATR performance under the EOC scenarios. The latter term includes a weighted pairwise similarity and a local linearity rule to capture the intrinsic manifold structure information. An overview of the LMF methods presented is shown in Table VIII.

*5) Deep Learning:*

During the last years, deep learning algorithms have been gaining large attention due to their high classification rates. An additional advantage is their end-to-end nature that includes all three major classification stages, i.e. feature detection, description, and matching. As described in the subsections, in the context of SAR ATR deep learning includes Convolutional Neural Networks, Restricted Boltzmann Machines, Recurrent Neural Networks, Stacked Autoencoders, and hybrid strategies combining deep learning methods. An overview of the methods presented is shown in Table IX.

TABLE VIII
LOW-RANK MATRIX FACTORIZATION

| Nº | Reference | Type | SAR chip size | Main features |
|----|-----------|------|---------------|---------------|
| 1 | Dang *et al*. [87] | NMF | 64x64 | $l_{1/2} - norm$ |
| 2 | Zhang *et al*. [88] | NMF | 64x64 | Gabor, PCA, and wavelet features, two-step classification: coarse using a voting scheme and fine using a Bayesian decision fusion |
| 3 | Dang *et al*. [15] | NMF | - | SRC $l_{1/2} - norm$ classification |
| 4 | Yu *et al*. [89] | TSVD | 64x64 | - |

### a) Convolutional Neural Networks (CNN)

Gao *et al*. [90] enhance the effectiveness of their CNN by employing a cost function during training that combines cross-entropy with the class separability information. The latter comprises of intra-class compactness and inter-class separability. Yue *et al*. [91] propose a CNN algorithm that combines both supervised and unsupervised training. Specifically, the unsupervised pipeline utilizes CNN to obtain the class probabilities of the unlabeled samples. Then the class probabilities are optimized via a hard thresholding process, which are then used to calculate the scatter matrices of the Linear Discriminant Analysis (LDA) method. Finally, the loss function of the supervised pipeline of the CNN is adapted by the scatter matrices created from the unsupervised pipeline. Chen *et al*. [6] and Wang *et al*. [92] propose a CNN that comprises only of sparsely connected layers (convolutional and pooling layers) neglecting the typical fully connected layers. This strategy reduces the algorithm's free parameters and thus overfitting due to limited training images. Their all-convolutional networks (A-ConvNets) is unarguably one of the top-performing SAR ATR techniques.

Wang *et al*. [93] realize the influence of speckle-noise to SAR ATR and reduce its impact by introducing a de-speckling CNN as a pre-processing step applied on the entire SAR dataset. The latter CNN comprises of five convolutional layers followed by a ReLU activation function while pooling layers are not used to preserve the feature map size through the entire network. Kwak *et al*. [94] also consider the influence of speckle-noise and propose a speckle noise-invariant CNN that employs a regularization term to minimize the feature variations caused by this type of noise. The regularization term minimizes the CNN features between the raw and the de-speckled image variant and contributes to the final ATR classification score in a weighted manner along with the raw image classification output.

One sub-class of CNN-based approaches involves CNN distillation to ultimately create a compact and shallow CNN that has the classification power of deeper CNNs. For example, Min *et al*. [95] extend the typical distillation scheme into the SAR ATR domain by adopting a student-teacher paradigm, where the teacher is a deep 18-layered CNN and the student a shallow two-layered ternary network (weights are -1, 0, 1). During training, the teacher's output vector after the SoftMax processing is used to train the student's CNN. The student's loss function is the sum of the cross-entropy between the student network output and the ground truth target label, and the cross-entropy between the teacher network output and the ground truth target label. The student CNN is fully trained when its loss function minimizes. Another approach towards lightweight designs is proposed by Zhang *et al*. [96], which adapts the A-ConvNets [6], [92] to exploit a pruning and a knowledge distillation scheme. Their method is a three-stage process where initially the A-ConvNet is forced to a lossy compression by pruning the feature maps. Then, the CNN's accuracy is recovered via a student-teacher type knowledge distillation concept, where the teacher is the original un-pruned A-ConvNet and the student its pruned variant. Trials demonstrated that the pruned A-ConvNets completely preserves its SAR ATR capability compared to the original A-ConvNets, despite being 68.7 times compressed and thus 2.5 faster to execute.

Another sub-class of CNN-based approaches exploits data augmentation to enhance CNN's classification performance by training it on more templates of various nuisances. Typical approaches create the additional templates manually, while more recent approaches rely on a Generative Adversarial Network (GAN) [97]–[99]. For example, Wagner [100] manually generates artificial training data that presents elastic distortion, rotation, and affine transformations, simulating a changing depression angle or an incorrectly estimated aspect angle, aiming at increasing the robustness against these nuisances. Similarly, Ding *et al*. [101] suggest augmenting the training dataset by generating images that include speckle noise, several translations, and pose variations. Experimental results demonstrate that indeed, data augmentation enhances the ATR classification rates. Yan [102] extends the training images by manually generating speckle-noisy samples at different signal-to-noise ratios (SNRs), multiresolution representations, and partially occluded images. Furthermore, the training samples are pre-processed to reduce the interferences between the original and the augmented training imagery by cropping the target to exclude background and then enhancing the image via a power function. The resulting image is then normalized and down sampled before being input to the CNN. Zhong and Ettinger [7] augment the CNN's training process by utilizing a pose-aware regression function that aims at minimizing the weighted sum between the CNN estimated target class score, and the pose difference between the predicted and the ground truth rotation of the target.

On the contrary, Cui *et al*. [103] suggest data augmentation via a Wasserstein Generative Adversarial Nets scheme [104] with a gradient penalty (WGAN-GP) [105]. WGAN extends the typical GAN by incorporating a Wasserstein distance metric to measure the distance between the distribution of generated data and the real data. WGAN-GP further extends a WGAN by applying the Lipschitz constraints instead of the weight clipping of WGAN, affording faster convergence, and generating higher quality samples than the original WGAN. Once the SAR imagery is augmented, it is then input to a typical CNN algorithm that exploits Leaky ReLU activations. Similarly, Sun et al. [106] establish an angular rotation GAN, where the attribute between the raw and the simulated SAR imagery is the target aspect angle. The synthetic images are then utilized to train a residual network (ResNet). Shi et al. [107] augment the training dataset by generating super-resolution SAR imagery via the GAN network. Up-sampling of the low-resolution images is created by introducing the two Pixel Shuffler layers after the residual block structure.

TABLE IX
DEEP LEARNING ALGORITHMS

| Nº | Reference | Type | SAR chip size | Pre-processing | Data augmentation | Main features |
|---|---|---|---|---|---|---|
| 1 | Zhang et al. [96] | CNN | 88x88 | - | - | CNN distillation of the A-ConvNets |
| 2 | Huang et al. [119] | CNN | 88x88 | - | Random cropping | two-stream CNN |
| 3 | Gao et al. [90] | CNN | 64x64 | - | translation | cost function for inter-class and intra-class separability enhancement |
| 4 | Yue et al. [91] | CNN | 64x64 | - | - | combination of supervised and unsupervised learning |
| 5 | Min et al. [95] | CNN | 128x128 | - | - | CNN distillation, ternary student network |
| 6 | Cui et al. [103] | CNN | 64x64 | - | GAN | Employs Leaky ReLU layers |
| 7 | Zhang et al. [121] | CNN and LSTM | 68x68 | - | Random crop for translation invariance | |
| 8 | Sun et al. [106] | CNN | 64x64 | $L_2$ normalization | Angular GAN | residual network algorithm |
| 9 | Pei et al. [122] | CNN | 90x90 | orientation alignment | multiple target views (4) | |
| 10 | Shang et al. [120] | CNN | 70x70 | - | Random clipping | Index table to store templates |
| 11 | Yan [102] | CNN | 72x72 | power function enhancement, normalization, down sampling | speckle noise, multi-resolution, and partial occlusion | manual data augmentation |
| 12 | Wang et al. [93] | CNN | 88x88 | noise de-speckling | | five convolutional layers |
| 13 | Zhong et al. [117] | CNN | 88x88 | - | Random cropping | transfer learning on AlexNet |
| 14 | Kwak et al. [94] | CNN | 88x88 | - | speckle noise | speckle noise-invariant CNN |
| 15 | Tian et al. [118] | CNN | 96x96 | | | intra-kernel relationship between layers |
| 16 | Kechagias-Stamatis [113] | CNN | 128x128 | - | - | multi-dimensional feature tensor of a shallow VGG layer |
| 17 | Zhong and Ettinger [7] | CNN | 128x128 | | Left-right image flipping | pose-aware regression function |
| 18 | Amrani and Jiang [116] | CNN | 128x128 | - | - | Concatenating three layers of various depths of the VGG network |
| 19 | Kechagias-Stamatis et al. [114] | CNN | 128x128 | - | - | clustering AlexNet layers, SVM for classification |
| 20 | Zhang et al. [123] | Bi-LSTM | 128x128 | - | - | Multiple views (4) |
| 21 | Chen et al. [6] | CNN | 88x88 | - | - | convolutional and pooling layers only |
| 22 | Huang et al. [124] | CNN and SAE | 128x128 | - | - | Transfer learning |
| 23 | Wang et al. [92] | CNN | 88x88 | - | - | convolutional and pooling layers only |
| 24 | Wagner [98] | CNN | 128x128 | | elastic Distortion, affine transformation, rotation | manual data augmentation, SVM for classification |
| 25 | Huang et al. [125] | RBM | 90x90 | - | - | biologically inspired model |
| 26 | Kang et al. [14] | SAE | 128x128 | - | - | feature fusion |
| 27 | Ding et al. [100] | CNN | 128x128 | | Pose, translation, Speckle noise | manual data augmentation |
| 28 | Deng et al. [13] | SAE | 128x128 | image normalization | - | - |

A number of approaches extend the usability of current CNNs trained for visual imagery into the SAR domain. These techniques commonly use the AlexNet [108], VGG [109], or ResNet [110], which are pre-trained on the ImageNet dataset

[111] that includes millions of optical images. The concept of this type of SAR ATR methods is to employ the knowledge of these CNNs on mid-level image representations such as the SAR imagery either by directly utilizing the raw weights of these pre-trained CNNs or by exploiting the transfer learning technique (TL) [112] to fine-tune the raw weights incorporating knowledge from the SAR data domain. For the former case, Kechagias-Stamatis [113] exploited directly the weights of the pre-trained VGG by clustering the VGG layers and extracting a multi-dimensional feature tensor of a shallow cluster. The reasoning is that shallow convolutional layers reveal the coarse object's features and thus the shallow feature maps of the pre-trained VGG are able to bridge the SAR – visual data domain gap. The target class is derived utilizing a Nearest-neighbor feature matching strategy exploiting the cosine similarity metric between the target and the training imagery feature tensors. Similarly, Kechagias-Stamatis et al. [114] cluster AlexNet and combine it with a multi-class Support Vector Machines (SVM) [115] for target classification. Amrani and Jiang [116] concatenate three layers of various depths of the VGG network

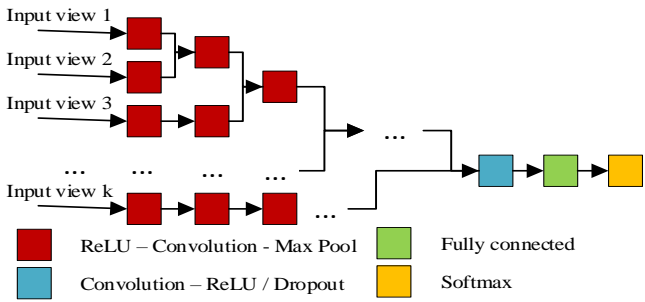

**Fig. 5.** The CNN progressively fuses the feature maps from each image aspect to the last layer of the previous fusion

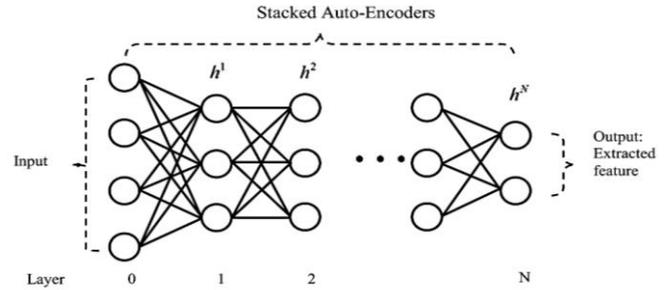

**Fig. 6.** Typical SAE algorithm. Several AE are stacked layer by layer to extract deep features. (h denotes the hidden features per layer)

features and create an index table containing these features along with the LDMLT similarity metric that measures the feature space of the training samples. Opposingly, Zhong *et al.* [117] applies the TL technique on the fifth convolutional layer of AlexNet, and also amend AlexNet with an additional convolutional and a pooling layer. Their results demonstrate that TL attains a higher classification performance compared to completely training the same CNN. Finally, they further enhance the performance by employing data augmentation by randomly cropping the SAR chip.

In contrast to typical CNN pipelines, Tian *et al.* [118] suggest a CNN where the weights of each layer are manipulated from the previous layers by establishing a strong intra-kernel relationship. Huang *et al.* [119] recommend a two-stream CNN to extract multi-level features, where the local stream uses global maximum pooling for significant local feature extraction and the global stream uses large-step convolution kernels for global feature extraction. Shang *et al.* [120] incorporate in their CNN algorithm an index table to store the training samples' spatial features and suggest the M-Net algorithm. Then for a query SAR image, this method calculates the spatial similarity between the query image features and the stored index table. To improve M-Net's convergence, the authors propose a two-staged training process. First, initial training is performed on solely the CNN module of the M-Net to initialize the CNN's parameters. Second, the initialized parameters from step one are then applied to the M-Net to update CNN's parameters and fine-tune the entire M-Net.

Opposing to the methods presented that employ a single target image, several solutions increase the classification rates by exploiting multiple aspects from the target. For example, Pei *et al.* [121] suggest a CNN topology that relies on multiple image views, where each aspect poses a distinct input. The CNN progressively fuses the feature maps from each image aspect to the last layer of the previous fusion, to ultimately classify the target. The fusion pipeline is presented in Fig. 5.

### b) Stacked Autoencoders (SAE)

Stacked Autoencoders (SAE) are sequentially placed Autoencoders (AE) (Fig. 6), where the latter is a neural network that comprises an encoder and a decoder. The encoder remaps the input data to a hidden representation $h$ according to

$$h = f(x) = s_f \left( W_x + b_h \right) \qquad (12)$$

where $s_f$ is a non-linear activation function, e.g. sigmoid or ReLU, $W$ is a weight matrix and $b_h$ a bias vector. Accordingly, the decoder remaps $h$ to a reconstructed version of the input data by minimizing a cost function. In terms of SAR ATR, only the encoding part on an AE is used, which is then input to a classification scheme, e.g. SVM. AEs are employed in a stacked fashion to increase descriptiveness.

Kang *et al.* [14] suggest a feature fusion process that involves 23 texture-based properties of the SAR image with features extracted via a Three-Patch Local Binary Pattern (TPLBP) process. The former features include properties such as area, bounding box, convex hull, etc., while the latter involves a set of classic Local Binary Patterns (LBP) distributed uniformly on a ring of radius *r* and one LBP is placed at the center of the ring. Features are then cascaded and input to an SAE. Deng *et al.* [13] reduce overfitting by modifying the reconstruction error of SAE by adding a Euclidean distance restriction for the hidden layer features followed by a *Dropout* layer. Other autoencoder based solutions are influenced by the human visual cortical system [122] or are combined with a Synergetic neural network concept [123].

### c) Restricted Boltzmann Machine (RBM)

Huang *et al.* [124] are inspired by the human cognition system and suggest a convolutional restricted Boltzmann machine (CRBM) algorithm that learns the episodic features which are then used to extract the semantic features of a SAR image. For the episodic feature extraction, a convolutional deep belief network is trained to segment the target's SAR reflection and its shadow. These regions are then input to the semantic feature extraction module that considers area, length, slope, and grey value calculation. The episodic features are further utilized to enhance ATR performance by estimating the aspect angle of the observed target. Training instances with aspect angles that exceed a $\pm 10°$ from the target aspect angle are discarded during the classification process.

### d) Hybrid Deep Learning Techniques

Huang *et al.* [125] suggest an ATR algorithm that combines the CNN and SAE concepts along with the Transfer Learning technique. The CNN pipeline consists of a classification and a reconstruction module that is used as a feedback to estimate the reconstruction loss. To avoid overfitting, the authors apply data augmentation and train the reconstruction module of the CNN pipeline with stacked convolutional auto-encoders (SCAE). Chen and Wang [17] also combine the concepts of SAE and CNN by involving a single layer-CNN that is trained on randomly sampled image patches utilizing an unsupervised sparse auto-encoder.

TABLE X
HYBRID REFLECTIVITY ATTRIBUTE-BASED METHODS

| Nº | Reference | Type | SAR chip size | Data augmentation | Main features |
|---|---|---|---|---|---|
| 1 | Zhang *et al.* [123] | features and deep learning | 128x128 | - | Gabor filters, LBP, LSTM |
| 2 | Lv and Liu [129] | CNN | 128x128 | ASC-based various orientations | - |
| 3 | Kechagias-Stamatis and Aouf [138] | CNN, SRC | 80x80 | - | decision fusion |
| 4 | Ding *et al.* [140] | ASC, SRC | 128x128 | - | feature fusion |
| 5 | Karine *et al.* [131] | features, SRC | 128x128 | - | decision fusion |
| 6 | Gishkori and Mulgrew [135] | Zernike moments, SRC | 96x96 | - | - |
| 7 | Ding *et al.* [136] | azimuth sensitivity, raw data | 128x128 | - | decision fusion |
| 8 | Zhang *et al.* [137] | Zernike moments, SRC | 70x70 | - | 2D slices of the SAR image along the amplitude direction |
| 9 | Sun *et al.* [133] | features, raw data | 65x65 | - | LC-KSVD learned dictionary |

Zhang *et al.* [126] combine a residual (ResNet) network and a bidirectional long short-term memory (BiLSTM) network. Fusing these two types of networks aims at extracting the target's scattering features from a single-aspect image (ResNet module) and extend their encoding over several adjacent multi-aspect images through the recurrent learning (BiLSTM module). The latter network is a variant of the popular Long Short-Term Memory (LSTM) layer. For an analysis of the operating principles of LSTM layers, the reader is referred to [127].

### 6) Hybrid reflectivity attribute-based methods

Table X presents the Hybrid reflectivity attribute-based algorithms. For example, Zhang *et al.* [128] combine the concepts of feature-based SAR ATR with deep learning. Specifically, the authors propose a multi-aspect algorithm where Gabor filters and three-patch local binary patterns extract comprehensive spatial features, which are then input to a multilayer perceptron network followed by a bidirectional LSTM recurrent neural network. Lv and Liu [129] propose a novel data augmentation algorithm where the ASC are extracted using a Sparse Representation (SR) scheme [52], [130]. ASCs are created with various $\theta_i$ parameters (Eq. (2)), the quality of which is evaluated based on a quality factor. Finally, the augmented dataset is used to train a CNN.

Spurred by the classification enhancement of decision fusion, current literature suggests various strategies of that type. For example, Karine *et al.* [131] combine the concepts of local features, saliency maps, and SRC. Specifically, given a SAR image, they create a saliency map inspired by the human visual system to establish a region of interest (ROI) type of area containing the target. Then the scale-invariant feature transform (SIFT) [132] (a 2D local keypoint detector and descriptor) is applied on the ROI and for each detected keypoint a feature vector is created. The latter ones are converted into a matrix that is used as a dictionary during the SRC process. Given that for this method each SAR image provides a 2D signature rather than a 1D, which is the norm for SRC, the authors employ a multitask SRC, where the reconstruction error originates from a block of atoms. Sun *et al.* [133] use the original SAR image and the SIFT features, where the former representation incorporates intensity information and the latter gradient information. Both representations are used jointly [134], while the learned dictionary relies on the LC-KSVD algorithm [77]. Gishkori and Mulgrew [135] propose a similar solution but utilize a set of rotation invariant Pseudo Zernike Moments (PZM) that are generated from the entire SAR image. Then the PZM are employed to create a dictionary, which is used during the SRC process. Ding *et al.* [136] suggest a decision-level fusion of a two-stream SRC pipeline. The first SRC utilizes the SAR imagery, while the second SRC the azimuth sensitivity image (ASI) which reveals the target sensitivity at an azimuth angle. Finally, the target identity is defined by comparing the classification scores of both SRC streams. Zhang *et al.* [137] generate a multilayer 2D slices of the SAR image along with the amplitude direction and then calculate a set of Zernike Moments (ZM) [2] for each slice. Finally, an SRC scheme is employed exploiting the ZM feature vector.

In [138] Kechagias-Stamatis and Aouf combine the strengths of SRC and CNN in a decision fusion scheme. For the SRC module, a novel adaptive elastic net type of optimization is proposed that balances the advantages of $l_1-norm$ and $l_2-norm$ depending on a Gaussian Mixture Model (GMM) [139] analysis of each SAR image. Regarding the CNN module, the authors suggest utilizing the AlexNet response map originating from a shallow layer, while target classification is relying on a multiclass SVM scheme. It should be noted that AlexNet is not retrained or any transfer learning is applied, but rather the pre-trained weights of AlexNet are employed. Finally, the target identity is determined based on a decision-level scheme that adaptively changes its fusion weights. In [140] Ding *et al.* combine the concepts of SRC and ASC by hierarchically fusing the global and local features of a SAR image. Hence, in the first stage, the Gaussian random projection [141] features are used as global features that are then employed for an SRC process. If the decision is reliable, i.e. the reconstruction error does not exceed a threshold, then the classification process terminates, while if not, local features based on the ASC are extracted. The target identity is defined by the template ASC closest to the target ASC based on the Hungarian algorithm.

### 7) Summary

Table IV to 10 give a summary of the reflectivity attributed-based methods. This category of techniques exploits the raw SAR image without remapping it into a different data domain.

a. Most recent papers involve deep learning structures with the vast majority relying on convolutional neural networks.

b. CNN-based architectures incorporate various schemes ranging from manual or GAN-based data augmentation to CNN distillation strategies, transfer learning and fully exploiting pre-trained state-of-the-art deep networks.

c. Sparse classification representation for SAR ATR is also commonly used involving various $l_p - norm$ optimizations.

d. Hybrid reflectivity attribute-based methods are also of great interest as these aim at fusing the strengths of the involving disciplines.

### B. Transform Based Methods

Opposing to the reflectivity attribute-based methods, the transform-based strategies remap the raw SAR imagery into a different data domain on which the ATR pipeline is applied.

### 1) Frequency domain

Dong *et al*. [142] remap the SAR imagery into the frequency domain utilizing the discrete Fourier transform (DFT). Then the authors apply the typical SRC employing an $l_1 - norm$.

### 2) Grassmann Manifold

The monogenic signal is a multidimensional generalization of the analytic signal, where the latter is the combination of the signal itself with its Hilbert transformed version, i.e. a submanifold of a Grassmannian manifold [143]. For a 1D signal $f(x)$, the analytic signal is

$$f_a(x) = f(x) + jH\{f\}(x) = A(x)\exp\{j\varphi(x)\} \qquad (13)$$

where $H\{\cdot\}$ is the Hilbert transform, $A(x)$ the local amplitude, and $\varphi(x)$ the local phase. In a polar-coordinate frame, $A(x)$ and $\varphi(x)$ are linked to a local quantitative measure and qualitative information of a signal respectively [144].

Accordingly, the monogenic signal employs a scalar-to-vector extension of Hilbert transform, the Riesz transform. Hence, a 2D monogenic signal is defined as:

$$f_m(z) = f(z) - (i, j)f_r(z) \qquad (14)$$

where $f(z)$ is the original signal, $i, j$ are imaginary units, $\{i, j, 1\}$ an orthonormal basis of $\mathfrak{R}^3$ and $f_r(z)$ its Riesz transformed version with

$$f_r(z) = h_r(z) * f(z) = \left[h_r^x(z) * f(z), h_r^y(z) * f(z)\right]^T \qquad (15)$$

where the Riesz kernel is

$$h_r(z) = -\frac{z}{2\pi|z|^3} \qquad (16)$$

and $h_r^x(z), h_r^y(z)$ are the first and second order Riesz kernels, respectively. However, the original signal has a finite length with periodic spectra and thus is infinitely extended via a band-pass filter,

$$f_m(z) = \left(f * h_{bp}\right)(z) - (i, j) * h_{bp} f_r(z) \qquad (17)$$

where $h_{bp}$ is a band-pass filter. In the context of SAR ATR, the monogenic signal of Eq. (14) is decomposed into its real part, the $i$-imaginary part, and the $j$-imaginary part, which are then further processed in a target recognition pipeline.

Dong *et al*. [145] generate the monogenic signal utilizing a log-Gabor filter. The monogenic signal is then employed in a dictionary learning strategy that jointly considers SRC and low-rank representation, with the former aiming at restricting the representation and the latter limiting the feasible set of dictionaries. Dictionary learning incorporates several class-specific sub-dictionaries, which are then combined into a single dictionary. Zhou *et al*. [146] reduce the impact of the information redundancy between the scales and the heterogeneous nature of the three components of the monogenic signal, by proposing a scale selection method based on a weighted multi-task joint sparse representation. The scale selection module relies on a Fisher score-based evaluation of each scale and monogenic component, where the scales and components with a high score are further utilized to produce a dictionary. The latter is then used in a multi-task joint sparse representation classification framework. Dong *et al*. [147] embed Grassmann manifolds into an implicit Reproducing Kernel Hilbert Space (RKHS) and create an overcomplete dictionary. In this technique, the Grassmann manifolds are build using steerable wavelet coefficients.

### C. Compressive Sensing (CS)

Compressing Sensing aims at recovering a signal that has been remapped from the originating domain to a domain where the signal is sparse, using a non-adaptive linear projection. Signal recovery is achieved via an $l_1$-norm optimization process. In the context of SAR ATR, Multitask CS (MtCS) [148] exploits the statistical correlation among multiple target views to recover the target's signature that is then used for target recognition under a compressive sensing scheme. Jin *et al*. [149] involve three features, namely the principal component analysis features (PCA) [150], elliptical Fourier descriptors (EFD) [151], and the azimuthal sensitivity image. The former two are extracted representing the intensity distribution and the target shape respectively, while the latter is generated and describes the electromagnetic scattering characteristics of the target. The three features are then utilized in an MtCS scheme to derive the target identity. Liu and Yang [152] suggest a feature-fusion scheme that combines the PCA, the non-linear extension of PCA namely the Kernel Principal Components Analysis (KPCA) [153], and the NMF features. Finally, these features are input to an MtCS scheme.

TABLE XI
TRANSFORM BASED METHODS

| Nº | Reference | Type | SAR chip size | Pre-processing | Main features |
|---|---|---|---|---|---|
| 1 | Dong *et al.* [147] | Grassmann manifold | 64x64 | - | employing steerable wavelet coefficients |
| 2 | Dong *et al.* [142] | frequency domain | 128x128 | - | SRC classification |
| 3 | Dong *et al.* [145] | Grassmann manifold | 128x128 | Log-Gabor-filtering | SRC |
| 4 | Zhou *et al.* [146] | Grassmann manifold | 64x64 | Log-Gabor-filtering | multi-task joint SRC |
| 5 | Jin *et al.* [149] | CS | 128x128 | - | PCA, elliptical Fourier descriptors and azimuthal sensitivity input to Multitask CS |
| 5 | Liu *et al.* [148] | CS | 128x128 | - | Multitask CS |
| 6 | Liu and Yang [152] | CS | 128x128 | - | PCA, KPCA, NMF features input to Multitask CS |

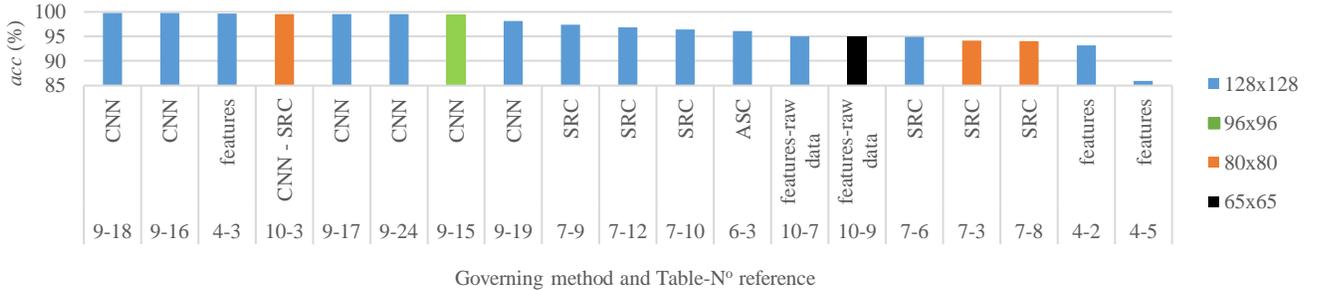

Fig. 7. 10-class SOC 1 SAR ATR performance (best seen in color)

*3.2.4 Summary:* Table XI gives a summary of the transformed-based methods. In contrast to the reflectivity attribute techniques, this category remaps the original SAR imagery into a different data domain.

a. Most recent papers involve Grassmann manifolds and compressive sensing.

b. Quantitatively, this category is less popular mainly due to the imbalance between the data domain remapping process complexity and the ATR performance gain over the methods of Section III-A (also see Section IV).

c. Sparse classification representation for SAR ATR is also commonly used involving various $l_p - norm$ optimizations.

d. Hybrid reflectivity attribute-based methods are also of great interest as these aim at fusing the strengths of the involving disciplines. The majority of the proposed ATR pipelines rely on recovering the sparsest solution among the available ones, either by employing an SRC or a CS strategy.

## IV. PERFORMANCE ANALYSIS

### A. Evaluation Criteria

For this work each method will be evaluated based on the accuracy (*acc*) metric:

$$acc = \frac{\Sigma True\ positive + \Sigma\ True\ negative}{\Sigma Total\ population} \quad (18)$$

However, despite the original MSTAR dataset offers a decoy object, in the vast majority of the relevant literature this object is omitted during the trials and therefore $\Sigma$ *True negative* = 0 .

As already noted (see Section 2), despite the MSTAR being a typical SAR ATR benchmarking dataset, it has quite a few deficiencies, one of which is the strong background correlation between the template and the target images imposing high true positive ATR rates despite the target absence from the scene [35], [36]. Thus, several algorithms crop or segment the target from the scene leading to several SAR image patch sizes ranging from 128x128 (the original image size) down to 64x64 pixels. Given that background correlation, and thus the SAR image size may indeed affect the overall ATR performance of a method, comparing methods that employ different SAR image sizes is not fair. Additionally, simply manipulating the SAR images to meet the algorithm's input image size constraints is still not an optimum choice as the entire parameter tuning of each method is linked to a specific image size. Finally, re-implementing and re-tuning each ATR technique is beyond the scope of this work. Therefore, for fairness in the following sections we highlight the SAR patch size involved.

### B. Evaluating SOC Subsets

The most commonly used 10-class SOC trial is SOC 1 (Table III). Its main feature is utilizing during training and testing all major model variants of BMP2 and T72. From Fig. 7 it is obvious that the governing methods are CNN and SRC-based, with an exception of method 4-3, i.e. Table IV method Nº 3. Feature-based methods are also available, but these tend to provide lower *acc* rates. CNN methods tend to be more appealing than their SRC counterparts, with the performance gain of the former being in the order of 4%. Specifically, for a SAR patch size of 128x128 pixels, methods 9-18 and 9-16 present the highest ATR performance demonstrating that partially exploiting structures of state-of-the-art deep networks used in the visual domain, i.e. the VGG for both these methods, can be highly beneficial. Interestingly, both 9-18 and 9-16 do not involve any Transfer Learning but employ the original VGG kernel weights. However, for smaller SAR images (96x96 pixels) pure CNN provides a relatively lower *acc* (method 9-15), while for an 80x80 pixel size, high ATR rates require more complex solutions such as a decision fusion scheme combining CNN and SRC (method 10-3). CNN-based methods offer a higher ATR performance than SRC because the latter down-samples the SAR image to create an overcomplete dictionary discarding some of the target's features, and ultimately diminishes the descriptiveness of the SRC method. Regarding the feature-based techniques, these are highly dependent on the originating feature extraction and description concept of each method. However, since each feature description method was originally designed for images in the visual domain, the performance of the feature-based methods for SAR ATR depends on the capability of each technique to handle equally well the SAR and the visual data domain.

The next most common 10-class SOC configuration used in the literature is SOC 2. Compared to SOC-1, this subset employs only a single target model for BMP2 and T72 rather than all available ones and is therefore considered of relatively lower complexity. Indeed, regardless of the ATR method and SAR image size used, the average *acc* on SOC-1 is 96.5%, while on SOC-2 99.1%.

Fig. 8 clearly shows that deep learning-based algorithms attain a minor performance gain over their counterparts. However, in contrast to the SOC-1 evaluation subset, here the *acc* difference is only 0.7% between the CNN and the SRC-based methods. Additionally, Fig. 8 shows that *acc* is less related to the SAR image size, as various image sizes are equally spread throughout the *acc* performance plot. Hence, the ATR performance is less affected by the background correlation as top-performing solutions involve SAR image patches with various background correlation levels. However, as stated earlier, the complexity of this target subset is lower allowing higher *acc* rates.

Regarding the 3-class SOC trials, SOC 1 is the most evaluated configuration (for details see Table III). This subset is similar to the 10-class SOC 1, in terms that for the BMP2, and T72 targets all major models are exploited both during training and testing. Thus, interestingly, the top-3 performing techniques are the same for both these SOC subsets, and indeed CNN-based algorithms are still performing relatively better than their competitors, with SRC and ASC to follow closely (Fig. 9). The CNN methods are more efficient as the fine details of the various training and testing models are properly encrypted by the convolutional layers. On the contrary, SRC-based pipelines down-sample the raw SAR image during the creation of the over-complete dictionary, discarding the fine details that differentiate the intra-class target variations. Similarly, ASC methods highlight the strong scatterers and discard the weak ones neglecting the fine details.

### C. Evaluating EOC Subsets

The vast majority of the SAR ATR papers also evaluate the effectiveness of the proposed methods against Extended Operation Conditions. The most referenced EOC scenarios are the ones presented in Table III that involve depression angle variation, target version variation, and several SAR image nuisances such as Gaussian noise, occlusion, and resolution variation at various levels.

The most referenced EOC trial investigates the robustness under depression angle variation where the ATR methods are trained on the 2S1, BRDM2 and the ZSU23/4 targets at 17° depression angle and tested on the same targets at 30° and 45° while some methods are also challenged at 15°. Thus, for the sake of uniformity, in this work, we present the performance of current methods only at 30° and 45° depression angles. In contrast to the SOC scenarios, Fig. 10 shows that CNN has only one representative solution, while the vast majority is SRC-based. This is because altering the depression angle creates on the target different electromagnetic reflection geometries that ultimately affect the SAR image. However, SRC during the over-complete dictionary creation smooths the local radar reflectivity parameters due to the depression angle variation and ultimately increases the robustness of the SRC methods. In contrast, the scattering centers are very much affected by the depression angle and thus techniques relying on ASC's are not suitable for this type of trial. Regarding the background correlation, from Fig. 10 it is evident that various SAR patch sizes may perform equally well. This is because the SRC, CS, and LMF methods smooth the effectiveness of the background correlation. Additionally, it is also important to note that despite the average *acc* at 30° is 98.15%, at 45° it is only 76.32% highlighting the poor robustness of the majority of the methods to such an extended depression angle variation. Despite that, method 7-11 still manages a high *acc* rate at 45° (91.72%) due to its hybrid-structured synthesis. A detailed *acc* performance is presented in Table XII.

Fig. 11 presents the ATR performance on additive Gaussian noise variation at SNR levels of $\{10, 5, 0, -5, -10\}dB$. Similar to the previous EOC scenarios, SRC is still an appealing method with ASC being more robust. However, fusing these two methods provides the highest *acc* value. However, it should be noted that the top-performing solutions have only a negligible performance difference.

Quite a few papers also evaluate the robustness on several resolution variations ranging from $0.7 \times 0.7\ m$ down to $0.3 \times 0.3\ m$ with a step size of $0.1 \times 0.1\ m$. The majority of the competitor techniques attain *acc* rates in the order of 95% (Fig. 12), with the 9-16 method (CNN-based) affording a 100% *acc*. It is worth noting that the vast majority of the evaluated methods utilize the entire SAR image (128 x 128 pixels).

Evaluating SAR ATR algorithms against version variants is also common. Fig. 13 presents the performance attained for scenarios involving training on the BMP2, T72, BTR60, and T62 targets, and testing on various BMP2 variants. For this type of scenario, a CNN type of solution manages the highest *acc*, with the majority of techniques being SRC-based. Interestingly, background correlation is of less influence. However, the latter is purely due to the testing conditions of each method, which is related to the SAR image input size and does not necessarily directly link with the overall performance of each method as each technique is only evaluated on a single SAR patch size.

TABLE XII
ACC ON EOC DEPRESSION ANGLE VARIATION

| | Patch size (pix.) | Table – N° reference | average | 30° | 45° |
|---|---|---|---|---|---|
| Li et al. | 80 | 7-11 | 95.21 | 98.71 | 91.72 |
| Ding et al. | 128 | 10-4 | 91.23 | 99.07 | 83.39 |
| Dong et al. | 64 | 11-1 | 87.95 | 98.20 | 77.70 |
| Jin et al. | 128 | 11-5 | 87.54 | 98.73 | 76.35 |
| Ren at al. | 64 | 7-5 | 87.05 | 98.29 | 75.82 |
| Chang and You | 128 | 7-6 | 87.05 | 99.07 | 75.03 |
| Zhang | 128 | 7-9 | 86.65 | 99.05 | 74.25 |
| Lv and Liu | 128 | 10-2 | 86.54 | 98.61 | 74.48 |
| Ren et al. | 64 | 7-4 | 86.52 | 97.02 | 76.03 |
| Ding and Wen. | 128 | 7-10 | 85.45 | 98.84 | 72.06 |
| Zhang and Liu. | 128 | 7-12 | 85.38 | 98.92 | 71.85 |
| Kechagias-Stamatis and Aouf | 80 | 10-3 | 85.24 | 99.61 | 70.87 |
| Ding et al. | 128 | 10-7 | 84.98 | 98.72 | 71.25 |
| Yu et al. | 64 | 8-4 | 84.53 | 91.29 | 77.78 |
| Ding and Wen. | 128 | 6-7 | 84.46 | 96.76 | 72.12 |
| Miao and Shan | 128 | 7-2 | 84.25 | 97.12 | 71.38 |
| Liu and Yang | 128 | 11-6 | 84.07 | 98.84 | 69.31 |
| Karine et al. | 128 | 10-5 | 52.45 | 68.58 | 36.32 |

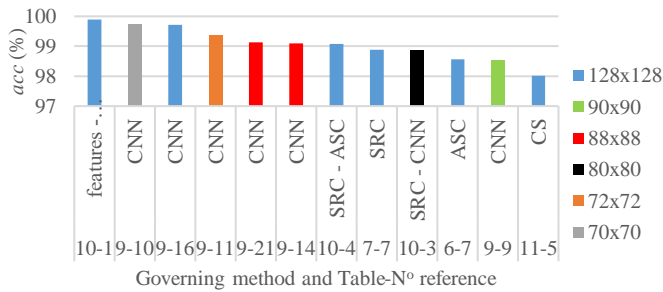

**Fig. 8.** Performance of 10-class SOC 2 (best seen in color)

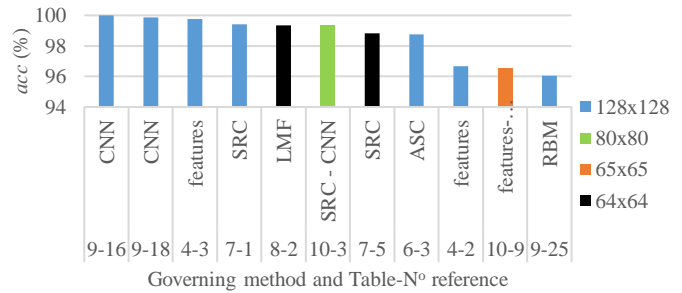

**Fig. 9.** Performance of 3-class SOC 1 (best seen in color)

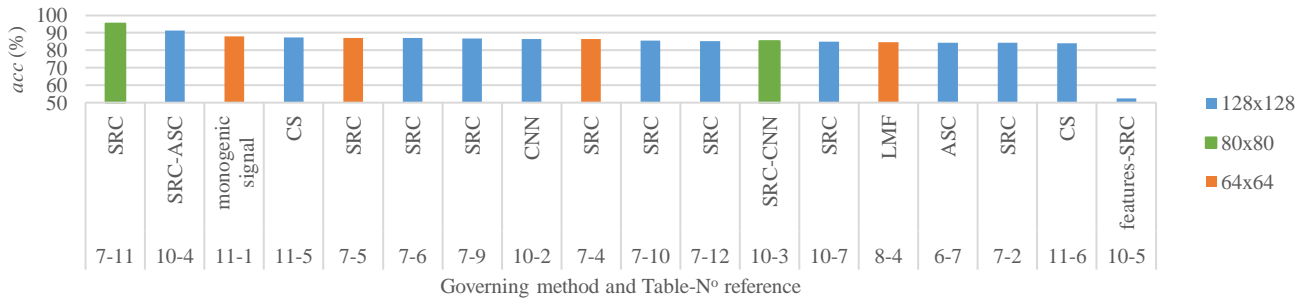

**Fig. 10.** EOC depression angle variation scenario – acc performance overview (best seen in color)

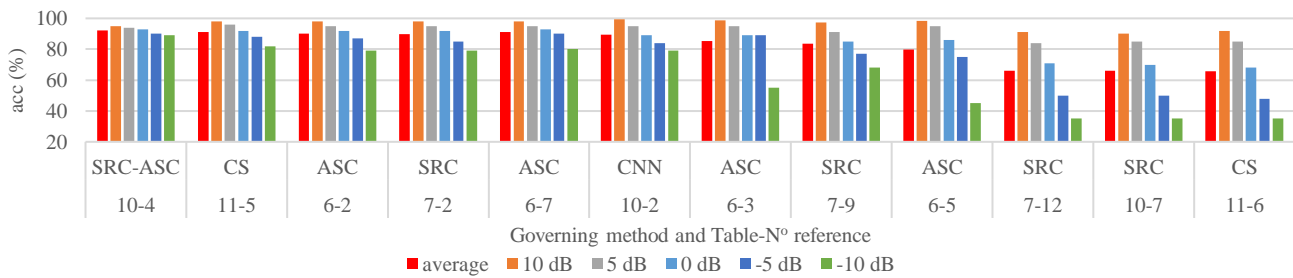

**Fig. 11.** EOC Gaussian noise level variation scenario (all methods utilize a 128x128 pixel SAR patch size – best seen in color)

Finally, occlusion variation has also been evaluated, with occlusion levels ranging from 10% up to 50% of the target with a step size of 10%. The performance plot of Fig. 14 highlights that most method categories are capable of an attractive SAR ATR performance.

### D. Summary

Table XIII presents an overview of the performance per method for the most commonly referenced SOC and EOC trials. From the analysis of this section, it can be succinctly interpreted that:

a. overall, deep learning-based methods are more robust on the SOC scenarios, while the SRC techniques on the EOC

scenarios. Despite CNN belong to deep learning methods, the CNN algorithms utilized for SAR ATR are relatively shallow with an average of four convolutional layers. This is related to the information content per pixel, which is much less than the corresponding one of the visual imagery.

b. CNN and SRC are appealing choices for both SOC and EOC scenarios, while the remaining techniques tend to present attractive ATR performance only on a subset of the scenarios. Indeed, ASC is robust only on EOC scenarios, while CS and hybrid reflectivity attribute pipelines on some of the SOC and EOC scenarios.

c. Transform-based methods and specifically frequency

TABLE XIII
OVERALL PERFORMANCE

| Scenario | most robust architectures |
|---|---|
| SOC 10-class | CNN, SRC |
| SOC 3-class | CNN, SRC |
| EOC depression angle variation | CNN, SRC, ASC, CS, Hybrid reflectivity attribute, LMF |
| EOC Gaussian noise level variation | CNN, SRC, ASC, CS, Hybrid reflectivity attribute |
| EOC resolution variation scenario | CNN, SRC, ASC, CS, Hybrid reflectivity attribute |
| EOC target version variation | SRC, CNN, LMF, ASC |
| EOC occlusion level variation | ASC, SRC, CNN, CS, features |

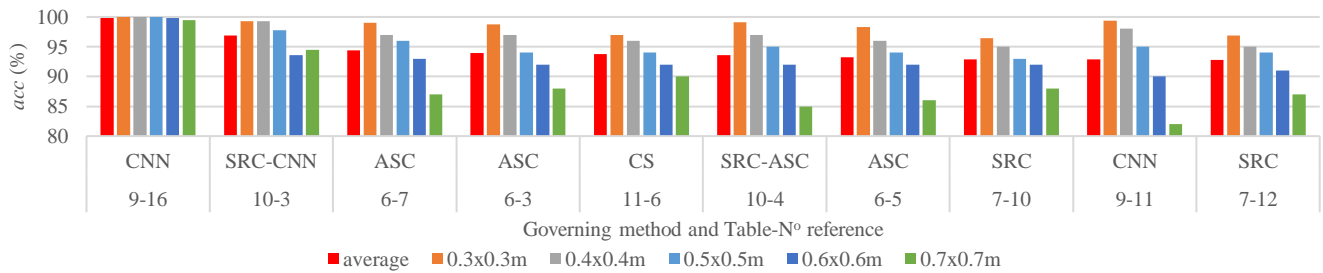

**Fig. 12.** EOC resolution variation scenario (method 10-3 employs an 80x80 SAR patch size, method 9-11 a 72x72, while the remaining methods a 128x128 - best seen in color)

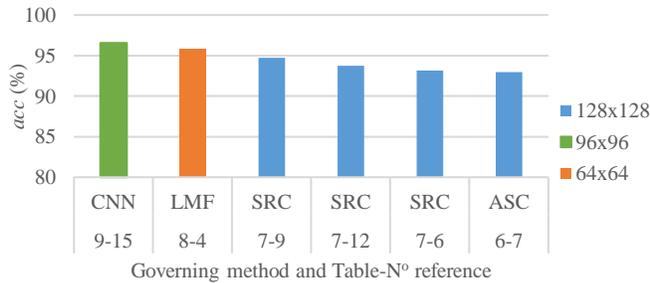

**Fig. 13.** Performance of EOC target version variation scenario (best seen in color)

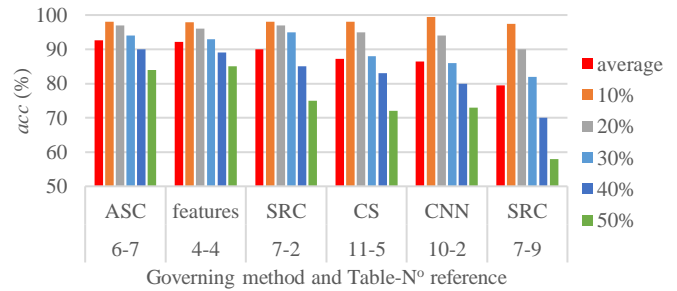

**Fig. 14.** EOC occlusion level variation scenario (method 4-4 uses an 88x88 SAR patch size - the remaining ones 128x128 - best seen in color)

domain and Grassmann manifolds have not been evaluated yet on the commonly used SOC and EOC trials.

d. Generally, the methods employing a 128x128 pixel SAR image tend to attain higher ATR rates. Though, comprehensive experiments are required to demonstrate if background correlation is indeed the major factor for such performance or the same SAR ATR pipelines are still capable of high *acc* rates at smaller SAR image patch sizes. It should be noted that simply feeding a SAR ATR algorithm with an image size different from the one originally designed for is not fair as the entire parameter tuning of the algorithm is based on a specific image size.

## V. CONCLUSION

This work presents a unique survey of current state-of-the-art SAR ATR algorithms that employ the MSTAR dataset. A comprehensive analysis, and comparative performance evaluation on most common SOC and EOC trials are also presented, highlighting the merits and demerits of each technique.

Future research directions may involve:

a. *Benchmark protocol*. Major deficiencies of the MSTAR database are the background correlation between the training and the testing images, and that all testing images are *a priory* known affecting the fine-tuning process of each method and biasing the true effectiveness of each technique. Therefore, creating a fixed benchmarking dataset originating from the MSTAR dataset that involves a common target segmentation scheme and several SOC and EOC testing images with publicly unknown class labels, would reveal the true effectiveness of each SAR ATR method. The latter could be conceptually related to the evaluation platform for visual odometry as presented in [154]. Additionally, standardizing the SOC and EOC trials would make direct performance comparison easier.

b. *Exploiting current synthetic SAR MSTAR-based datasets*. Recently the Synthetic and Measured Paired Labelled Experiment (SAMPLE) dataset has been proposed, as a complementary SAR ATR dataset [155]. Future work could utilize this dataset, which however may present the same background constraints as the original MSTAR dataset.

c. *Explainable deep learning*. As presented in Section 5, each class of methods has its strengths and weaknesses, with CNN-based techniques being more robust for the SOC scenarios and SRC on the EOC scenarios. Spurred by that, future work should include further research on explainable deep learning structures (XAI) to realize the underlying mechanisms of such networks and ultimately exploit that knowledge to further enhance the performance of CNN and expand their appealing performance to the EOC scenarios.

d. *Data augmentation*. Proven the performance enhancement of data augmentation, it would be interesting to evaluate the effectiveness of several data augmentation methods in conjunction with current ATR methods such as ASC and SRC.

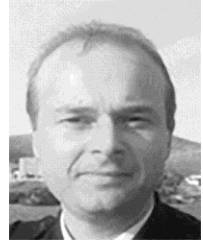

**Odysseas Kechagias-Stamatis** received the MSc degree in Guided Weapon Systems and the PhD degree in 3D ATR for missile platforms from Cranfield University, U.K. in 2011 and 2017, respectively. His research interests include 2D/3D object recognition and tracking, data fusion and autonomy of systems.

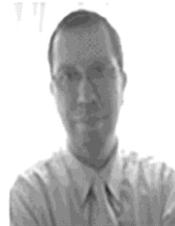

**Nabil Aouf** is currently professor of Robotics and Autonomous Systems and also leads the Robotics, Autonomy and Machine Intelligence (RAMI) Group at the Department of Electrical and Electronic Engineering, University of London, London, UK. He has authored over 100 publications in high caliber in his domains of interest. His research interests are aerospace and defense systems, information fusion and vision systems, guidance and navigation, tracking, and control and autonomy of systems. He is an Associate Editor of the International Journal of Computational Intelligence in Control.